\documentclass[10pt,twocolumn,letterpaper]{article}

\usepackage{cvpr}
\usepackage{times}
\usepackage{epsfig}
\usepackage{graphicx}
\usepackage{amsmath}
\usepackage{amssymb}
\usepackage{authblk}

\usepackage{algorithm}
\usepackage{algpseudocode}
\usepackage{multirow}
\usepackage{color}

\usepackage{tikz}
\usepackage{pgfplots}
\usepackage{paralist}
\usepackage{enumitem}
\usepackage{xcolor}

\definecolor{mygray}{HTML}{C6C6C6}
\definecolor{myblue}{HTML}{1A6EAA}

\newcommand{\del}[1]{{\color{mygray}#1}}

\newcommand{\comment}[1]{{\color{orange}#1}}

\renewcommand{\del}[1]{}

\renewcommand{\comment}[1]{}

\newcommand{\conv}[1]{\mathrm{conv_{#1\!\times\!#1}}}

\makeatletter
\def\BState{\State\hskip-\ALG@thistlm}
\makeatother

\usepackage[breaklinks=true,bookmarks=false]{hyperref}

\cvprfinalcopy 

\def\cvprPaperID{2343} 

\ifcvprfinal\fi
\begin{document}

\title{EAST: An Efficient and Accurate Scene Text Detector}

\author{Xinyu Zhou}
\author{Cong Yao}
\author{He Wen}
\author{Yuzhi Wang}
\author{Shuchang Zhou}
\author{Weiran He}
\author{Jiajun Liang}
\affil{Megvii Technology Inc., Beijing, China \\ \{zxy, yaocong, wenhe, wangyuzhi, zsc, hwr, liangjiajun\}@megvii.com}

\maketitle
\thispagestyle{empty}

\begin{abstract}

Previous approaches for scene text detection have already achieved promising performances across various benchmarks. However, they usually fall short when dealing with challenging scenarios, even when equipped with deep neural network models, because the overall performance is determined by the interplay of multiple stages and components in the pipelines. In this work, we propose a simple yet powerful pipeline that yields fast and accurate text detection in natural scenes. The pipeline directly predicts words or text lines of arbitrary orientations and quadrilateral shapes in full images, eliminating unnecessary intermediate steps (e.g., candidate aggregation and word partitioning), with a single neural network. The simplicity of our pipeline allows concentrating efforts on designing loss functions and neural network architecture. Experiments on standard datasets including ICDAR 2015, COCO-Text and MSRA-TD500 demonstrate that the proposed algorithm significantly outperforms state-of-the-art methods in terms of both accuracy and efficiency. On the ICDAR 2015 dataset, the proposed algorithm achieves an F-score of 0.7820 at 13.2fps at 720p resolution.
\vspace{-4mm}

\end{abstract}

\section{Introduction} \label{Sec:Introduction}

Recently, extracting and understanding textual information embodied in natural scenes have become increasingly important and popular, which is evidenced by the unprecedented large numbers of participants of the ICDAR series contests ~\cite{Ref:Shahab2011, Ref:Karatzas2013, Ref:Karatzas2015} and the launch of the TRAIT 2016 evaluation by NIST~\cite{Ref:TRAIT2015}.

Text detection, as a prerequisite of the subsequent processes, plays a critical role in the whole procedure of textual information extraction and understanding. Previous text detection approaches~\cite{Ref:Busta2015, Ref:Tian2015, Ref:Jaderberg2016B, gupta2016synthetic, Ref:Zhang2016} have already obtained promising performances on various benchmarks in this field. The core of text detection is the design of features to distinguish text from backgrounds. Traditionally, features are manually designed~\cite{Ref:Epshtein2010, Ref:Neumann2010, Ref:Yao2012, Ref:Huang2013, Ref:Neumann2012, Ref:Yin2014} to capture the properties of scene text, while in deep learning based methods~\cite{Ref:Coates2011, Ref:Jaderberg2014, Ref:Huang2014, Ref:Jaderberg2016B, gupta2016synthetic, Ref:Zhang2016} effective features are directly learned from training data.

\begin{figure}[!t]
\centering\begin{tikzpicture}
\pgfplotsset{compat=1.8}
\pgfplotsset{every tick label/.append style={font=\footnotesize}}
\pgfplotsset{grid style={dotted,gray}}
\begin{axis}[
xmin=0, xmax=17,
ymin=0.49, ymax=0.85,
grid=both,
axis lines=middle,
line width=.8pt,
width=1.1\linewidth,
height=.7\linewidth,
xlabel={\small Speed (FPS)},
ylabel={\small F-score}
]

\addplot [only marks, draw=black, fill=blue, mark size=3pt]
table {%
x     y
+7.14 +0.6085
+1.61 +0.6477
+0.476 +0.5358
};
\addplot [only marks, draw=black, fill=red, mark size=3pt]
table {%
x           y
+13.245e+00 +0.7820
+16.779e+00 +0.7571
};

\node at (axis cs:7.3,0.58)[
  anchor=base west,
  text=black,
  rotate=0.0,
  align=left
]{Tian~\etal~\cite{tian2016detecting}\\(0.609@7.14fps)};
\node at (axis cs:1.3,0.6677)[
  anchor=base west,
  text=black,
  rotate=0.0,
  align=left
]{Yao~\etal~\cite{yao2016scene}\\(0.648@1.61fps)};

\node at (axis cs:0.6,0.505)[
  anchor=base west,
  text=black,
  rotate=0.0,
  align=left
]{Zhang~\etal~\cite{Ref:Zhang2016}\\(0.532@0.476fps)};

\node at (axis cs:13.0,0.78)[
  anchor=base east,
  text=black,
  rotate=0.0,
  align=right
]{\bf Ours+PVANet2x\\(0.782@13.2fps)};

\node at (axis cs:16.8,0.69)[
  anchor=base east,
  text=black,
  rotate=0.0,
  align=right
]{\bf Ours+PVANet\\(0.757@16.8fps)};
\end{axis}

\end{tikzpicture}
\vspace{-2mm}
\caption{Performance versus speed on ICDAR 2015~\cite{Ref:Karatzas2015} text localization challenge. As can be seen, our algorithm significantly surpasses competitors in accuracy, whilst running very fast. The specifications of hardware used are listed in Tab.~\ref{tab:speed-comp}.}
\label{fig:perf-vs-speed}
\vspace{-6mm}
\end{figure}
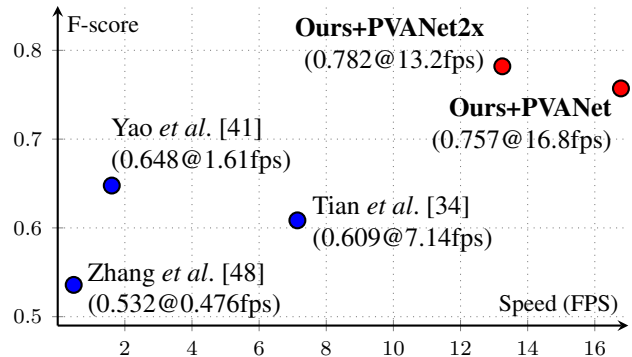

However, existing methods, either conventional or deep
neural network based, mostly consist of several stages and
components, which are probably sub-optimal and time-consuming. Therefore, the accuracy and efficiency of such methods are still far from satisfactory.

\begin{figure*}
\vspace{-2mm}
\centering\includegraphics[width=0.8\linewidth]{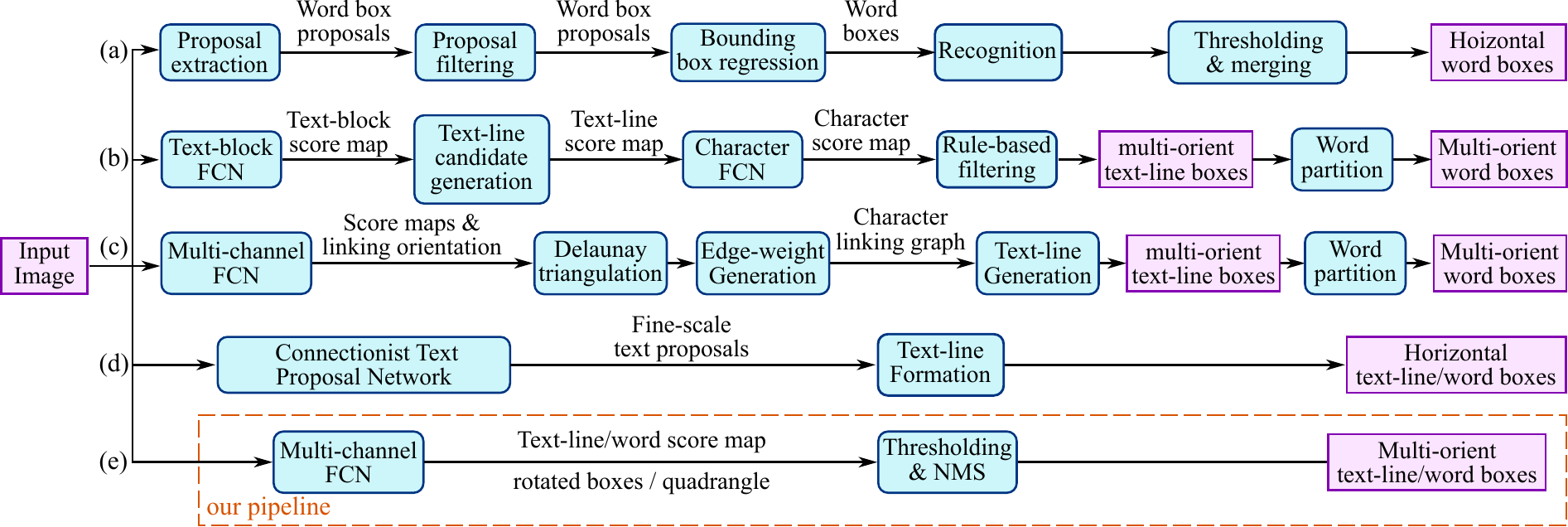}
\caption{Comparison of pipelines of several recent works on scene text detection:
	(a) Horizontal word detection and recognition pipeline proposed by Jaderberg~\etal~\cite{Ref:Jaderberg2016B};
	(b) Multi-orient text detection pipeline proposed by Zhang~\etal~\cite{Ref:Zhang2016};
	(c) Multi-orient text detection pipeline proposed by Yao~\etal~\cite{yao2016scene};
	(d) Horizontal text detection using CTPN, proposed by Tian~\etal~\cite{tian2016detecting};
	(e) Our pipeline, which eliminates most intermediate steps, consists of only two stages and is much simpler than previous solutions.
} \label{fig:pipeline}
\vspace{-3mm}
\end{figure*}

In this paper, we propose a fast and accurate scene text detection pipeline that has only two stages. The pipeline utilizes a fully convolutional network~(FCN) model that directly produces word or text-line level predictions, excluding redundant and slow intermediate steps. The produced text predictions, which can be either rotated rectangles or quadrangles, are sent to Non-Maximum Suppression to yield final results. Compared with existing methods, the proposed algorithm achieves significantly enhanced performance, while running much faster, according to the qualitative and quantitative experiments on standard benchmarks.

Specifically, the proposed algorithm achieves an F-score of 0.7820 on ICDAR 2015~\cite{Ref:Karatzas2015} (0.8072 when tested in multi-scale), 0.7608 on MSRA-TD500~\cite{Ref:Yao2012} and 0.3945 on COCO-Text~\cite{Ref:Veit2016}, outperforming previous state-of-the-art algorithms in performance while taking much less time on average (13.2fps at 720p resolution on a Titan-X GPU for our best performing model, 16.8fps for our fastest model).

The contributions of this work are three-fold:
\begin{itemize}[itemsep=0pt,parsep=0pt,topsep=0pt]
\item We propose a scene text detection method that consists of two stages: a Fully Convolutional Network and an NMS merging stage. The FCN directly produces text regions, excluding redundant and time-consuming intermediate steps.
\item The pipeline is flexible to produce either word level or line level predictions, whose geometric shapes can be rotated boxes or quadrangles, depending on specific applications.
\item The proposed algorithm significantly outperforms state-of-the-art methods in both accuracy and speed.
\end{itemize}

\section{Related Work}

Scene text detection and recognition have been active research topics in computer vision for a long period of time. Numerous inspiring ideas and effective approaches~\cite{Ref:Epshtein2010, Ref:Neumann2010, Ref:Neumann2012, Ref:Mishra2012, Ref:Novikova2012, Ref:Weinman2013, Ref:Huang2014, Ref:Jaderberg2016B, gupta2016synthetic,yao2016scene, Ref:Yao2014, Ref:Shi2016} have been investigated. Comprehensive reviews and detailed analyses can be found in survey papers~\cite{Ref:Zhu2016, Ref:Uchida2014, Ref:Ye2014}. This section will focus on works that are mostly relevant to the proposed algorithm.

Conventional approaches rely on manually designed features. Stroke Width Transform (SWT)~\cite{Ref:Epshtein2010} and Maximally Stable Extremal Regions (MSER)~\cite{Ref:Neumann2010, Ref:Neumann2012} based methods generally seek character candidates via edge detection or extremal region extraction. Zhang~\etal~\cite{Ref:Zhang2015} made use of the local symmetry property of text and designed various features for text region detection. FASText~\cite{Ref:Busta2015} is a fast text detection system that adapted and modified the well-known FAST key point detector for stroke extraction. However, these methods fall behind of those based on deep neural networks, in terms of both accuracy and adaptability, especially when dealing with challenging scenarios, such as low resolution and geometric distortion.

Recently, the area of scene text detection has entered a new era that deep neural network based algorithms~\cite{Ref:Huang2014, Ref:Jaderberg2014, Ref:Zhang2016, gupta2016synthetic} have gradually become the mainstream.
Huang~\etal~\cite{Ref:Huang2014} first found candidates using MSER and then employed a deep convolutional network as a strong classifier to prune false positives.
The method of Jaderberg~\etal~\cite{Ref:Jaderberg2014} scanned the image in a sliding-window fashion and produced a dense heatmap for each scale with a convolutional neural network model. Later, Jaderberg~\etal~\cite{Ref:Jaderberg2016B} employed both a CNN and an ACF to hunt word candidates and further refined them using regression. Tian~\etal~\cite{tian2016detecting} developed vertical anchors and constructed a CNN-RNN joint model to detect horizontal text lines. Different from these methods, Zhang~\etal~\cite{Ref:Zhang2016} proposed to utilize FCN~\cite{Ref:Long2015} for heatmap generation and to use component projection for orientation estimation. These methods obtained excellent performance on standard benchmarks.  However, as illustrated in Fig.~\ref{fig:pipeline}(a-d), they mostly consist of multiple stages and components, such as false positive removal by post filtering, candidate aggregation, line formation and word partition. The multitude of stages and components may require exhaustive tuning, leading to sub-optimal performance, and add to processing time of the whole pipeline.

In this paper, we devise a deep FCN-based pipeline that directly targets the final goal of text detection: word or text-line level detection.
As depicted in Fig.~\ref{fig:pipeline}(e), the model abandons unnecessary intermediate components and steps, and allows for end-to-end training and optimization.
The resultant system, equipped with a single, light-weighted neural network, surpasses all previous methods by an obvious margin in both performance and speed.

\section{Methodology}

The key component of the proposed algorithm is a neural network model, which is trained to directly predict the existence of text instances and their geometries from full images. The model is a fully-convolutional neural network adapted for text detection that outputs dense per-pixel predictions of words or text lines. This eliminates intermediate steps such as candidate proposal, text region formation and word partition. The {post-processing steps only include} thresholding and NMS on predicted geometric shapes. The detector is named as \textbf{EAST}, since it is an \textbf{E}fficient and \textbf{A}ccuracy \textbf{S}cene \textbf{T}ext detection pipeline.

\subsection{Pipeline} \label{pipeline}

A high-level overview of our pipeline is illustrated in Fig.~\ref{fig:pipeline}(e). The algorithm follows the general design of DenseBox~\cite{huang2015densebox}, in which an image is fed into the FCN and multiple channels {of pixel-level text score map and geometry are generated}.

One of the predicted channels is a score map whose pixel values are in the range of $[0, 1]$. The remaining channels represent geometries that encloses the word from the view of each pixel. The score stands for the confidence of the geometry shape predicted at the same location.

We have experimented with two geometry shapes for text regions, rotated box~(RBOX) and quadrangle~(QUAD), and designed different loss functions for each geometry. Thresholding is then applied to each predicted region, where the geometries whose scores are over the predefined threshold is considered valid and saved for later non-maximum-suppression. Results after NMS are considered the final output of the pipeline.

\subsection{Network Design}
\label{model}
\begin{figure}
\centering\includegraphics[width=.75\linewidth]{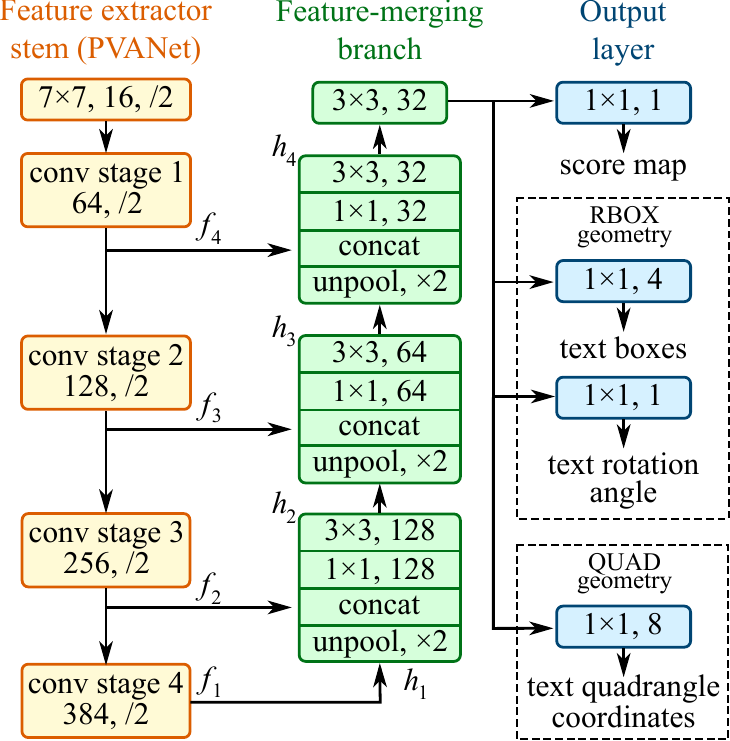}
\caption{Structure of our text detection FCN. }
\label{fig:network-structure}
\vspace{-0.55cm}
\end{figure}

Several factors must be taken into account when designing neural networks for text detection. Since the sizes of word regions, as shown in Fig.~\ref{fig:qualitative}, vary tremendously, determining the existence of large words would require features from late-stage of a neural network, while predicting accurate geometry enclosing a small word regions need low-level information in early stages. Therefore the network must use features from different levels to fulfill these requirements.
HyperNet~\cite{kong2016hypernet} meets these conditions on features maps, but merging a large number of channels on large feature maps would significantly increase the computation overhead for later stages.

In remedy of this, we adopt the idea from \textit{U-shape}~\cite{ronneberger2015u} to merge feature maps gradually, while keeping the up-sampling branches small. Together we end up with a network that can both utilize different levels of features and keep a small computation cost.

A schematic view of our model is depicted in Fig.~\ref{fig:network-structure}. The model can be decomposed in to three parts: feature extractor \textit{stem}, feature-merging \textit{branch} and output layer.

The~\textit{stem} can be a convolutional network pre-trained on \makebox{ImageNet~\cite{Ref:Deng2009}} dataset, with interleaving convolution and pooling layers.
Four levels of feature maps, denoted as $f_i$, are extracted from the stem, whose sizes are
$\frac{1}{32}$, $\frac{1}{16}$, $\frac{1}{8}$ and $\frac{1}{4}$ of the input image, respectively.
In Fig.~\ref{fig:network-structure}, PVANet~\cite{KimKH2016arXivPVANET} is depicted. In our experiments, we also adopted the well-known
VGG16~\cite{simonyan2014very} model, where feature maps after pooling-2 to pooling-5 are extracted.

In the feature-merging branch, we gradually merge them:
\begin{align}
g_i &= \begin{cases}
  \mathrm{unpool}(h_i) & \text{if}\quad i \le 3 \\
  \conv{3}(h_i) & \text{if}\quad i = 4
\end{cases} \\
h_i &= \begin{cases}
  f_i & \text{if}\quad i = 1 \\
  \conv{3}(\conv{1}([g_{i-1}; f_i])) & \text{otherwise}
\end{cases}
\end{align}
where $g_i$ is the merge base, and $h_i$ is the merged feature map, and the operator $[\cdot; \cdot]$ represents concatenation along the channel axis.
In each merging stage, the feature map from the last stage is first fed to an unpooling layer to double its size, and then concatenated with the current feature map. Next, a $\conv{1}$ bottleneck~\cite{he2015deep} cuts down the number of channels and reduces computation, followed by a $\conv{3}$ that fuses the information to finally produce the output of this merging stage.  Following the last merging stage, a $\conv{3}$ layer produces the final feature map of the merging branch and feed it to the output layer.

The number of output channels for each convolution is shown in Fig.~\ref{fig:network-structure}.
We keep the number of channels for convolutions in \textit{branch} small, which adds only a fraction of computation overhead over the
\textit{stem}, making the network computation-efficient. The final output layer contains several $\conv{1}$ operations to project 32 channels of feature maps into 1 channel of score map $F_\text{s}$ and a multi-channel geometry
map $F_\text{g}$. The geometry output can be either one of RBOX or QUAD, summarized in Tab.~\ref{tab:output-geometry}

For RBOX, the geometry is represented by 4 channels of axis-aligned bounding box~(AABB) $\mathbf{R}$ and 1 channel rotation angle $\theta$.
The formulation of $\mathbf{R}$ is the same as that in \cite{huang2015densebox}, where the 4 channels represents 4 distances from the pixel
location to the top, right, bottom, left boundaries of the rectangle respectively.

For QUAD $\mathbf{Q}$, we use 8 numbers to denote the coordinate shift from four corner vertices $\{p_i \,|\, i\!\!\in\!\!\{1, 2, 3, 4\}\}$ of the quadrangle to the pixel location.
As each distance offset contains two numbers $(\Delta x_i, \Delta y_i)$, the geometry output contains 8 channels.

\begin{table}
\footnotesize
\begin{center}
\begin{tabular}{|c|c|c|}
\hline
Geometry & channels & description \\
\hline\hline
AABB & 4 & $\mathbf{G} = \mathbf{R} = \{d_i | i \in \{1,2,3,4\}\}$\\\hline
RBOX & 5 & $\mathbf{G} = \{\mathbf{R}, \theta\}$ \\\hline
QUAD & 8 & $\mathbf{G} = \mathbf{Q} = \{(\Delta x_i, \Delta y_i) | i \in \{1,2,3,4\}\}$ \\\hline
\end{tabular}
\end{center}
\caption{Output geometry design} \label{tab:output-geometry}
\vspace{-4mm}
\end{table}

\subsection{Label Generation}
\label{gtgen}
\subsubsection{Score Map Generation for Quadrangle}

\begin{figure}
\centering\includegraphics[width=\linewidth]{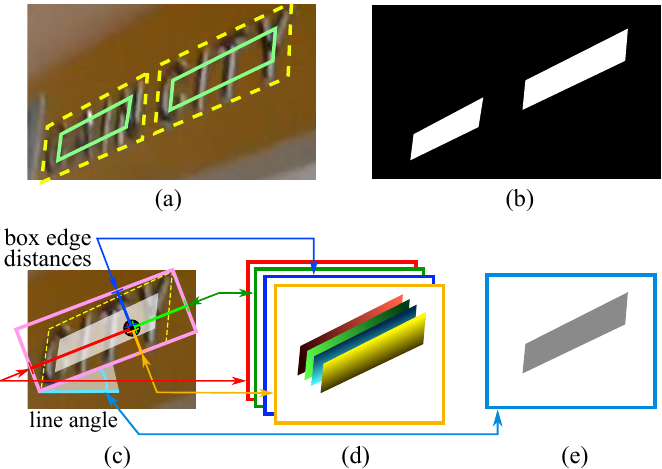}
\caption{Label generation process:
  (a) Text quadrangle (yellow dashed) and the shrunk quadrangle (green solid);
  (b) Text score map;
  (c) RBOX geometry map generation; (d) 4 channels of distances of each pixel to rectangle boundaries; (e) Rotation angle.
} \label{fig:gt-generation}
\vspace{-4mm}
\end{figure}

Without loss of generality, we only consider the case where the geometry is a quadrangle.
The positive area of the quadrangle on the score map is designed to be roughly a shrunk version of the original one, illustrated in Fig.~\ref{fig:gt-generation}~(a).

For a quadrangle $\mathbf{Q} = \{p_i | i \in \{1,2,3,4\}\}$, where $p_i = \{x_i, y_i\}$ are vertices on the quadrangle in clockwise order. To shrink $\mathbf{Q}$, we first compute
a \textit{reference length} $r_i$ for each vertex $p_i$ as
\begin{equation}
\begin{aligned}
  r_i =\min ( & \mathrm{D}(p_i, p_{(i \bmod 4) + 1}), \\
                & \mathrm{D}(p_i, p_{((i + 2) \bmod 4) + 1})) \\
\end{aligned}
\end{equation}
where $\mathrm{D}(p_i, p_j)$ is the $L_2$ distance between $p_i$ and $p_j$.

We first shrink the two longer edges of a quadrangle, and then the two shorter ones. For each pair of two opposing edges, we determine the ``longer'' pair by comparing the mean of their lengths.
For each edge $\langle p_i, p_{(i \bmod 4) + 1}\rangle$, we shrink it by moving its two endpoints inward along the edge by $0.3 r_i$ and $0.3 r_{(i \bmod 4) + 1}$ respectively.

\subsubsection{Geometry Map Generation}
As discussed in Sec.~\ref{model}, the geometry map is either one of RBOX or QUAD. The generation process for RBOX is illustrated in Fig.~\ref{fig:gt-generation}~(c-e).

For those datasets whose text regions are annotated in QUAD style (\eg, ICDAR 2015), we first generate a rotated rectangle that covers
the region with minimal area.
Then for each pixel which has positive score, we calculate its distances to the 4 boundaries of the text box, and put them to the 4 channels of
RBOX ground truth.
For the QUAD ground truth, the value of each pixel with positive score in the 8-channel geometry map is its coordinate
shift from the 4 vertices of the quadrangle.

\subsection{Loss Functions} \label{loss}

The loss can be formulated as
\begin{equation}
  L = L_\text{s} + \lambda_\text{g} L_\text{g}
\end{equation}
where $L_\text{s}$ and $L_\text{g}$ represents the losses for the score map and the geometry, respectively, and $\lambda_\text{g}$ weighs the importance between two losses. In our experiment, we set $\lambda_\text{g}$ to 1.

\subsubsection{Loss for Score Map}
In most state-of-the-art detection pipelines, training images are carefully processed by balanced sampling and hard negative mining to
tackle with the imbalanced distribution of target objects~\cite{huang2015densebox,ren2015faster}.  Doing so would potentially improve the network performance.  However, using such techniques inevitably introduces a non-differentiable stage and more parameters to tune and a more complicated pipeline, which contradicts our design principle.

To facilitate a simpler training procedure, we use class-balanced cross-entropy introduced in~\cite{xie2015holistically}, given by
\newcommand\bxent{\operatorname{balanced-xent}}
\begin{equation}
\begin{aligned}
    L_\text{s} & = \bxent(\mathbf{\hat{Y}}, \mathbf{Y^*}) \\
        & = -\beta \mathbf{Y^*} \log \mathbf{\hat{Y}} - (1 - \beta) (1 - \mathbf{Y^*}) \log (1 - \mathbf{\hat{Y}})
\end{aligned}
\end{equation}
where $\mathbf{\hat{Y}} = F_\text{s}$ is the prediction of the score map, and $\mathbf{Y^*}$ is the ground truth. The
parameter $\beta$ is the balancing factor between positive and negative samples, given by
\begin{equation}
\begin{aligned}
  \beta = 1 - \dfrac{\sum_{y^* \in \mathbf{Y^*}} y^*}{|\mathbf{Y^*}|}.
\end{aligned}
\end{equation}

This balanced cross-entropy loss is first adopted in text detection by Yao~\etal~\cite{yao2016scene} as the objective function for score map
prediction. We find it works well in practice.

\subsubsection{Loss for Geometries}
One challenge for text detection is that the sizes of text in natural scene images vary tremendously. Directly using L1 or L2 loss for
regression would guide the loss bias towards larger and longer text regions.  As we need to generate accurate
text geometry prediction for both large and small text regions, the regression loss should be scale-invariant. Therefore, we adopt the IoU loss
in the AABB part of RBOX regression, and a scale-normalized smoothed-L1 loss for QUAD regression.

\paragraph{RBOX}

For the AABB part, we adopt IoU loss in~\cite{yu2016unitbox}, since it is invariant against objects of different scales.
\begin{equation}
\begin{aligned}
  L_\text{AABB} & = -\log \mathrm{IoU}(\mathbf{\hat{R}}, \mathbf{R^*})  = -\log \dfrac{|\mathbf{\hat{R}} \cap \mathbf{R^*}|}{|\mathbf{\hat{R}} \cup \mathbf{R^*}|}
\end{aligned}
\end{equation}
where $\hat{\mathbf{R}}$ represents the predicted AABB geometry and $\mathbf{R}^*$ is its corresponding ground truth. It is easy to see that the width and height of the intersected rectangle $|\mathbf{\hat{R}} \cap \mathbf{R^*}|$ are
\begin{equation}
\begin{aligned}
w_{\mathbf{i}} & = \min (\hat{d}_2, d^*_2) + \min(\hat{d}_4, d^*_4) \\
h_{\mathbf{i}} & = \min (\hat{d}_1, d^*_1) + \min(\hat{d}_3, d^*_3) \\
\end{aligned}
\end{equation}
where $d_1$, $d_2$, $d_3$ and $d_4$ represents the distance from a pixel to the top, right, bottom and left boundary of its corresponding rectangle, respectively. The union area is given by
\begin{equation}
|\mathbf{\hat{R}} \cup \mathbf{R^*}| = |\mathbf{\hat{R}}| + |\mathbf{R^*}| - |\mathbf{\hat{R}} \cap \mathbf{R^*}|.
\end{equation}
Therefore, both the intersection/union area can be computed easily. Next, the loss of rotation angle is computed as
\begin{equation}
L_{\theta}(\hat{\theta}, \theta^*) = 1 - \cos (\hat{\theta} - \theta^*).
\end{equation}
where $\hat{\theta}$ is the prediction to the rotation angle and $\theta^*$ represents the ground truth. Finally, the overall geometry loss is the weighted sum of AABB loss and angle loss, given by
\begin{equation}
L_\text{g}= L_\text{AABB} + \lambda_{\theta}L_{\theta}.
\end{equation}
Where $\lambda_{\theta}$ is set to $10$ in our experiments.

Note that we compute $L_\text{AABB}$ regardless of rotation angle. This can be seen as an approximation of quadrangle IoU when the angle is perfectly predicted. Although it is not the case during training, it could still impose the correct gradient for the network to learn to predict $\mathbf{\hat{R}}$.

\paragraph{QUAD}

We extend the smoothed-L1 loss proposed in~\cite{girshick2015fast} by adding an extra normalization term designed for word quadrangles,
which is typically longer in one direction.  Let all coordinate values of $\mathbf{Q}$ be an ordered set
\begin{equation}
  \mathrm{C}_{\mathbf{Q}} = \{x_1, y_1, x_2, y_2, \dots, x_4, y_4\}
\end{equation}
then the loss can be written as
\begin{equation}
\begin{aligned}
    L_\text{g}& = L_{\text{QUAD}}(\mathbf{\hat{\mathbf{Q}}}, \mathbf{Q^*}) \\
    & = \min_{\tilde{\mathbf{Q}} \in P_{\mathbf{Q}^*}}
        \sum_{\substack{c_i \in \mathrm{C}_{\mathbf{Q}}, \\ \tilde{c}_i \in \mathrm{C}_{\mathbf{\tilde{Q}}}}}
        \frac{\mathrm{smoothed}_{L1} (c_i - \tilde{c}_i)}{8\times{N}_{\mathbf{Q}^*}}  \\
\end{aligned}
\end{equation}
where the normalization term $\mathrm{N}_{\mathbf{Q}^*}$ is the shorted edge length of the quadrangle, given by
\begin{equation}
    {N}_{\mathbf{Q}^*} = \min_{i=1}^4 D(p_i, p_{(i \bmod 4)+1}),
\end{equation}
and $P_{\mathbf{Q}}$ is the set of all equivalent quadrangles of $\mathbf{Q}^*$ with different vertices ordering. This ordering permutation
is required since the annotations of quadrangles in the public training datasets are inconsistent.

\subsection{Training} \label{training}

The network is trained end-to-end using ADAM~\cite{kingma2014adam} optimizer.
To speed up learning, we uniformly sample 512x512 crops from images to form a minibatch of size 24. Learning rate of ADAM starts from 1e-3, decays to one-tenth every 27300 minibatches, and stops at 1e-5. The network is trained until performance stops improving.

\subsection{Locality-Aware NMS}

To form the final results, the geometries survived after thresholding should be merged by NMS. A na\"{i}ve NMS algorithm runs in $O(n^2)$，
where $n$ is the number of candidate geometries, which is unacceptable as we are facing tens of thousands of geometries from dense predictions.

Under the assumption that the geometries from nearby pixels tend to be highly correlated, we proposed to merge the geometries row by row, and while merging geometries in the same row, we will iteratively merge the geometry currently encountered with the last merged one. This improved technique runs in $O(n)$ in best scenarios\footnote{Consider the case that only a single text line appears the image. In such case, all geometries will be highly overlapped if the network is sufficiently powerful}. Even though its worst case is the same as the na\"{i}ve one, as long as the locality assumption holds, the algorithm runs sufficiently fast in practice. The procedure is summarized in Algorithm~\ref{algo:nms-hier}

It is worth mentioning that, in $\textproc{WeightedMerge}(g, p)$, the coordinates of merged quadrangle are weight-averaged by the scores of two given quadrangles. To be specific, if $a = \textproc{WeightedMerge}(g, p)$, then $a_i = V(g) g_i + V(p) p_i$ and $V(a) = V(g) + V(p)$, where $a_i$ is one of the coordinates of $a$ subscripted by $i$, and $V(a)$ is the score of geometry $a$.

In fact, there is a subtle difference that we are "averaging" rather than "selecting" geometries, as in a standard NMS procedure will do, acting as a voting mechanism, which in turn introduces a stabilization effect when feeding videos. Nonetheless, we still adopt the word "NMS" for functional description.

\iftrue
\begin{algorithm}[t]
\caption{Locality-Aware NMS}\label{algo:nms-hier}
\begin{algorithmic}[1]
\Function{NMSLocality}{$geometries$}

\State $S \gets \varnothing$,~~$p \gets \varnothing$
\For {$g \in geometries$ in row first order}
  \If {$p \neq \varnothing \land \textproc{ShouldMerge}(g, p)$}
    \State $p \gets \textproc{WeightedMerge}(g, p)$
    \Else
    \If {$p \neq \varnothing$}
      \State $S \gets S \cup \{p\}$
        \EndIf
        \State $p \gets g$
    \EndIf
\EndFor

\If {$p \neq \varnothing$}
  \State $S \gets S \cup \{p\}$
\EndIf

\State \Return $\textproc{StandardNMS}(S)$

\EndFunction
\end{algorithmic}
\end{algorithm}
\vspace{-2mm}
\fi

\section{Experiments}

To compare the proposed algorithm with existing methods, we conducted qualitative and quantitative experiments on three public benchmarks: ICDAR2015, COCO-Text and MSRA-TD500. 

\subsection{Benchmark Datasets}

\textbf{ICDAR 2015} is used in Challenge 4 of ICDAR 2015 Robust Reading Competition~\cite{Ref:Karatzas2015}. It includes a total of 1500 pictures, 1000 of which are used for training and the remaining are for testing. The text regions are annotated by 4 vertices of the quadrangle, corresponding to the QUAD geometry in this paper. We also generate RBOX output by fitting a rotated rectangle which has the minimum area. These images are taken by Google Glass in an incidental way. Therefore text in the scene can be in arbitrary orientations, or suffer from motion blur and low resolution. We also used the 229 training images from ICDAR 2013.

\textbf{COCO-Text}~\cite{Ref:Veit2016} is the largest text detection dataset to date. It reuses the images from MS-COCO dataset~\cite{lin2014microsoft}. A total of 63,686 images are annotated, in which 43,686 are chosen to be the training set and the rest 20,000 for testing. Word regions are annotated in the form of axis-aligned bounding box (AABB), which is a special case of RBOX. For this dataset, we set angle $\theta$ to zero. We use the same data processing and test method as in ICDAR 2015.

\textbf{MSRA-TD500}~\cite{Ref:Yao2012} is a dataset comprises of 300 training images and 200 test images. Text regions are of arbitrary orientations and annotated at sentence level. Different from the other datasets, it contains text in both English and Chinese. The text regions are annotated in RBOX format. Since the number of training images is too few to learn a deep model, we also harness 400 images from HUST-TR400 dataset~\cite{yao2014unified} as training data.

\subsection{Base Networks}

As except for COCO-Text, all text detection datasets are relatively small compared to the datasets for general object detection\cite{krizhevsky2012imagenet,lin2014microsoft}, therefore if a single network is adopted for all the benchmarks, it may suffer from either over-fitting or under-fitting. We experimented with three different base networks, with different output geometries, on all the datasets to evaluate the proposed framework. These networks are summarized in Tab.~\ref{tab:base-models}.

\textbf{VGG16}~\cite{simonyan2014very} is widely used as base network in many tasks~\cite{ren2015faster,xie2015holistically} to support subsequent task-specific fine-tuning, including text detection~\cite{tian2016detecting,Ref:Zhang2016,zhong2016deeptext,gupta2016synthetic}. There are two drawbacks of this network: (1). The receptive field for this network is small. Each pixel in output of conv5\_3 only has a receptive field of 196. (2). It is a rather large network.

\textbf{PVANET} is a light weight network introduced in ~\cite{KimKH2016arXivPVANET}, aiming as a substitution of the feature extractor in Faster-RCNN~\cite{ren2015faster} framework. Since it is too small for GPU to fully utilizes computation parallelism, we also adopt~\textit{PVANET2x} that doubles the channels of the original PVANET, exploiting more computation parallelism while running slightly slower than PVANET. This is detailed in Sec.~\ref{sec:speed}. The receptive field of the output of the last convolution layer is 809, which is much larger than VGG16.

The models are pre-trained on the ImageNet dataset~\cite{krizhevsky2012imagenet}.

\begin{table}
\begin{center}
\begin{tabular}{|c|c|}
\hline
\textbf{Network} & \textbf{Description} \\
\hline\hline
PVANET~\cite{KimKH2016arXivPVANET} & small and fast model \\\hline
PVANET2x~\cite{KimKH2016arXivPVANET} & PVANET with 2x number of channels \\\hline
VGG16~\cite{simonyan2014very} & commonly used model \\\hline
\end{tabular}
\end{center}
\caption{Base Models} \label{tab:base-models}
\vspace{-4mm}
\end{table}

\subsection{Qualitative Results}

\begin{figure*}
\begin{center}

\begin{tabular}{ccc}
	\begin{minipage}{0.3\linewidth}
		\includegraphics[width=\linewidth]{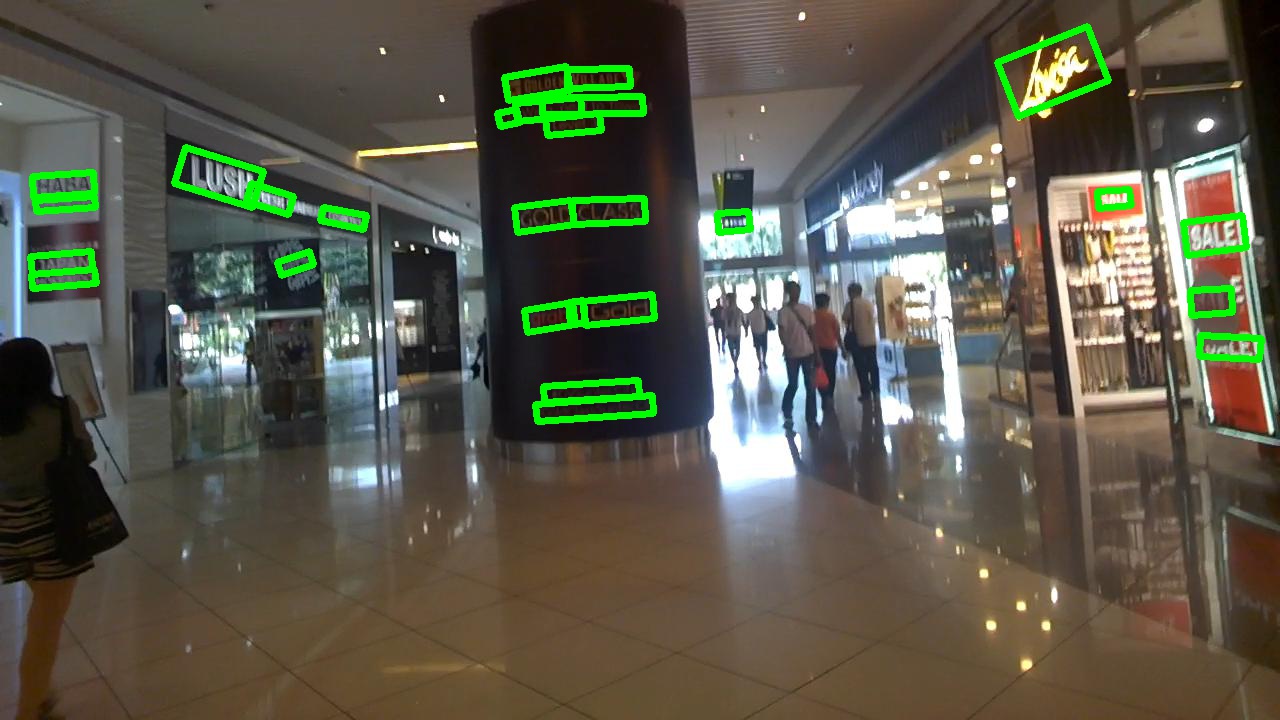}

		\vspace{2mm}

		\includegraphics[width=\linewidth]{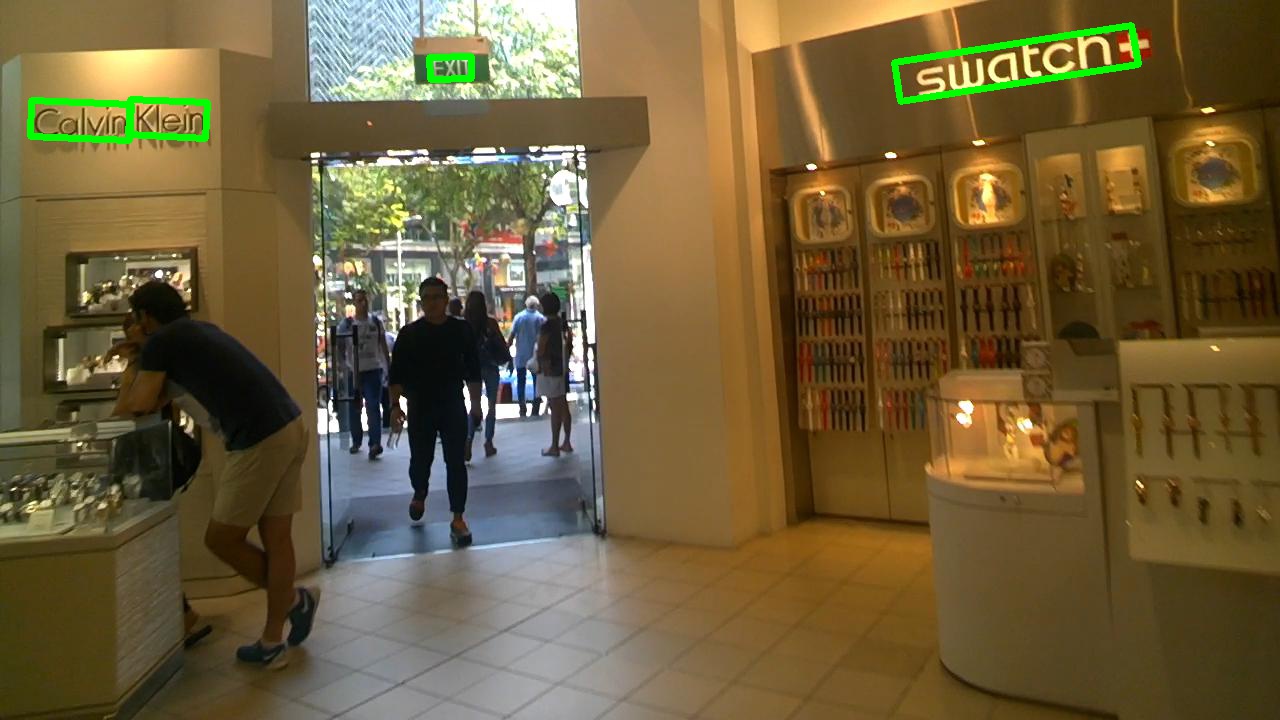}

        \vspace{1mm}

        \centering{(a)}
	\end{minipage}
&
	\begin{minipage}{0.22\linewidth}
		\includegraphics[width=\linewidth]{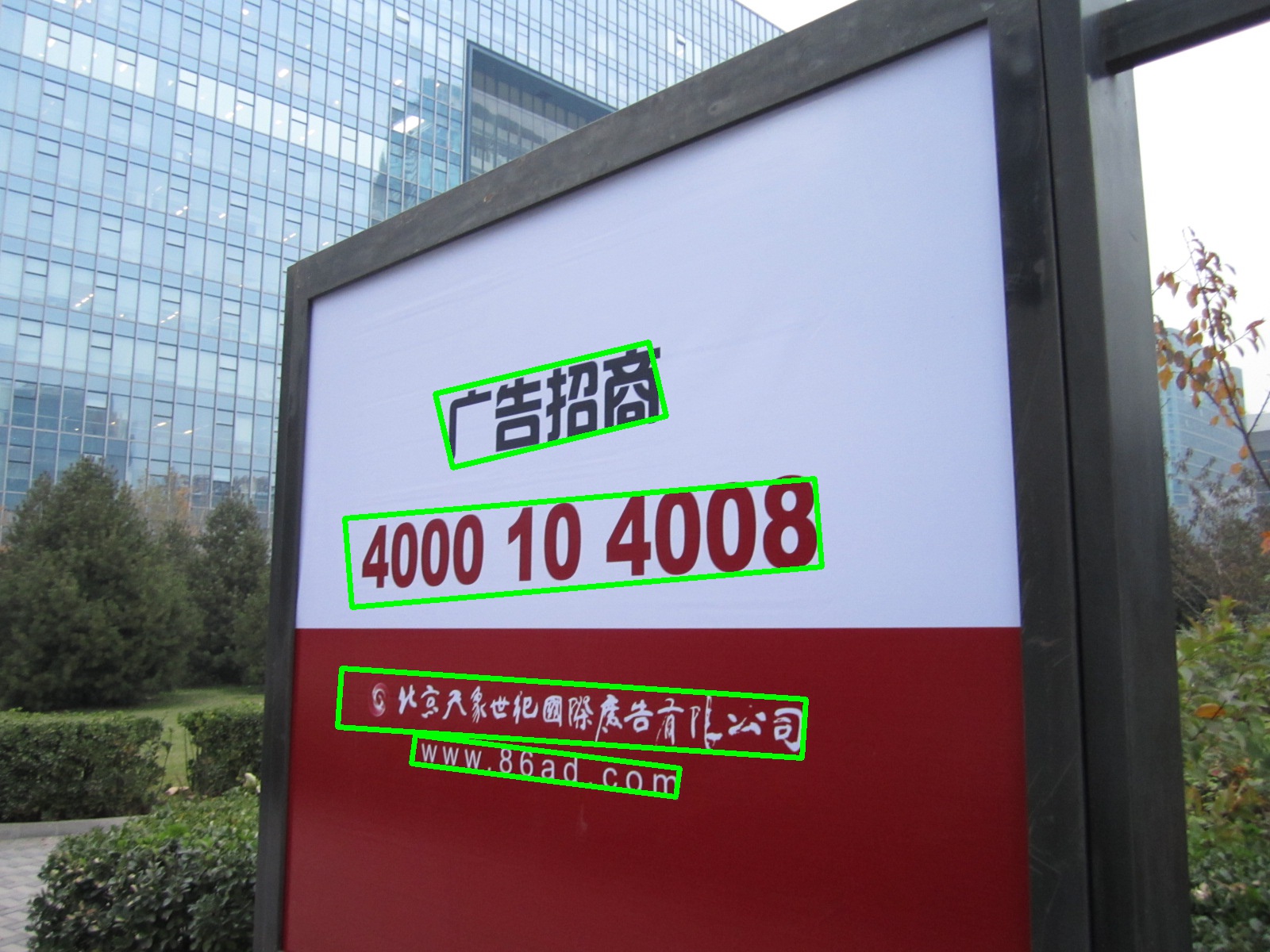}

		\vspace{3mm}

		\includegraphics[width=\linewidth]{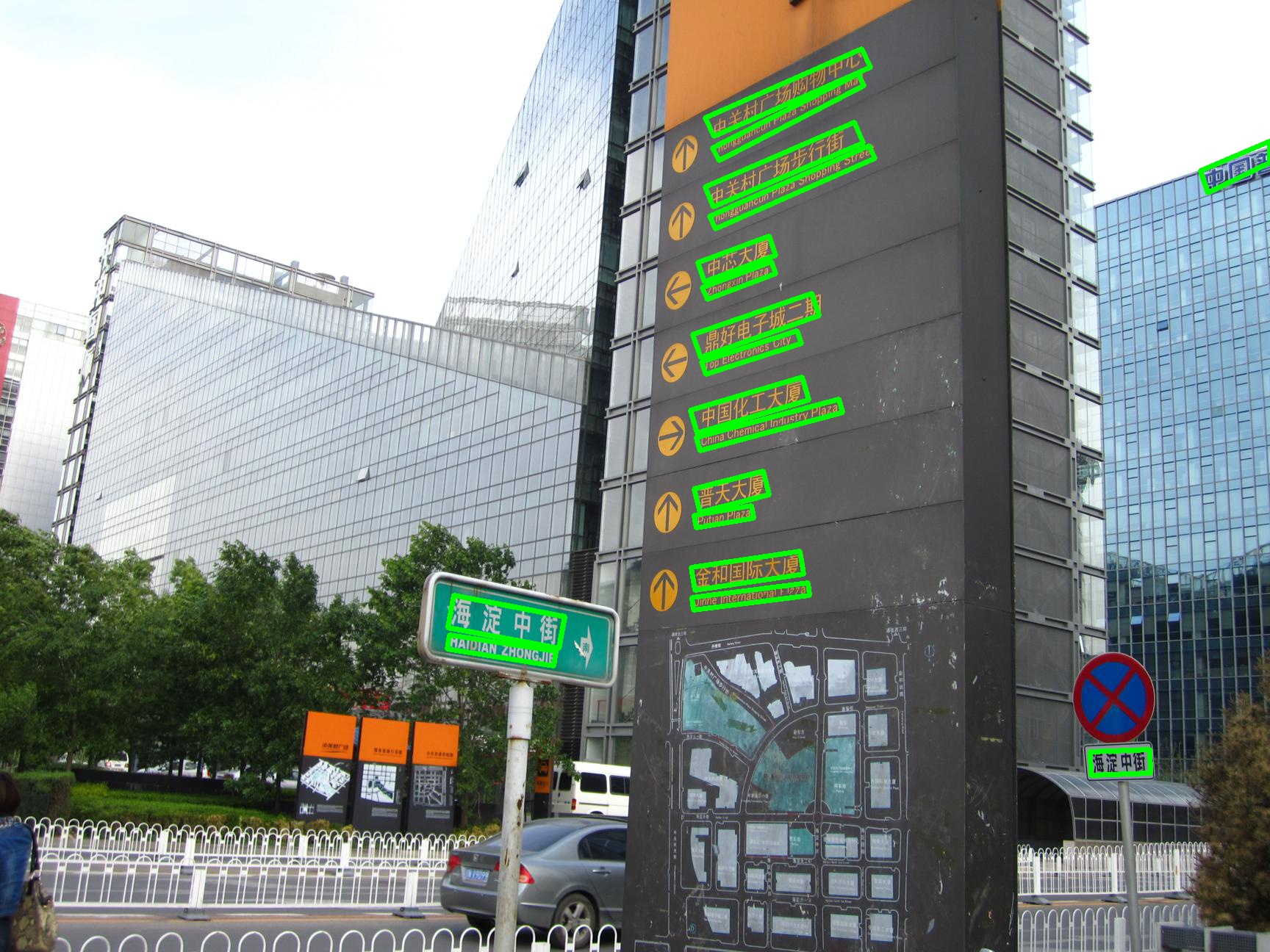}

        \vspace{1mm}

        \centering{(b)}
	\end{minipage}
&
	\begin{minipage}{0.22\linewidth}
		\includegraphics[width=\linewidth]{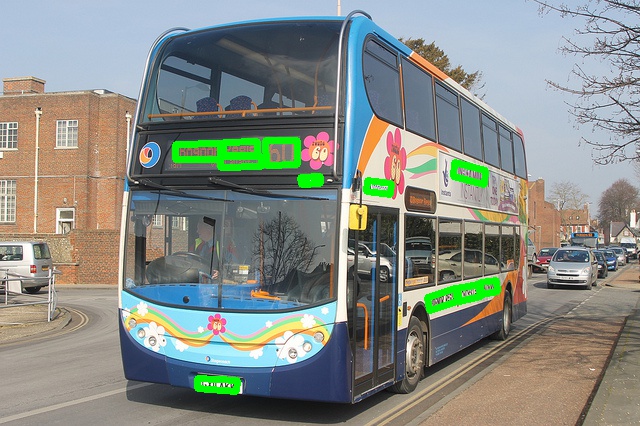}

		\vspace{5mm}

		\includegraphics[width=\linewidth]{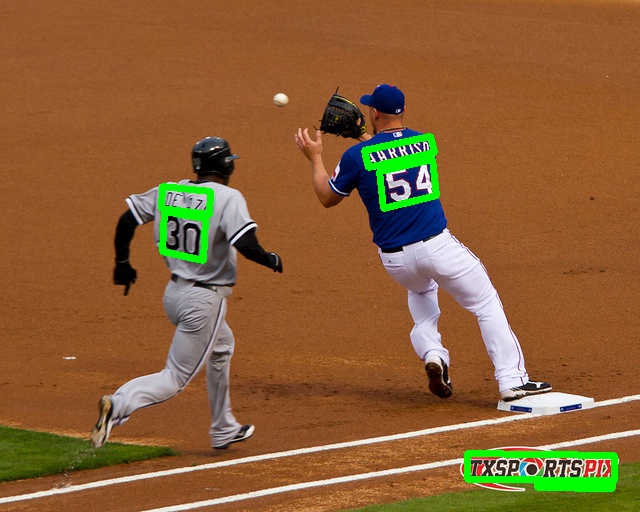}

        \vspace{1mm}

        \centering{(c)}
	\end{minipage} \\

\end{tabular}

\end{center}
\caption{Qualitative results of the proposed algorithm. (a) ICDAR 2015. (b) MSRA-TD500. (c) COCO-Text.} \label{fig:qualitative}
\vspace{-3mm}
\end{figure*}

\begin{figure}[!t]
\begin{center}
\begin{tabular}{ccc}
	\begin{minipage}{0.45\linewidth}
		\includegraphics[width=\linewidth]{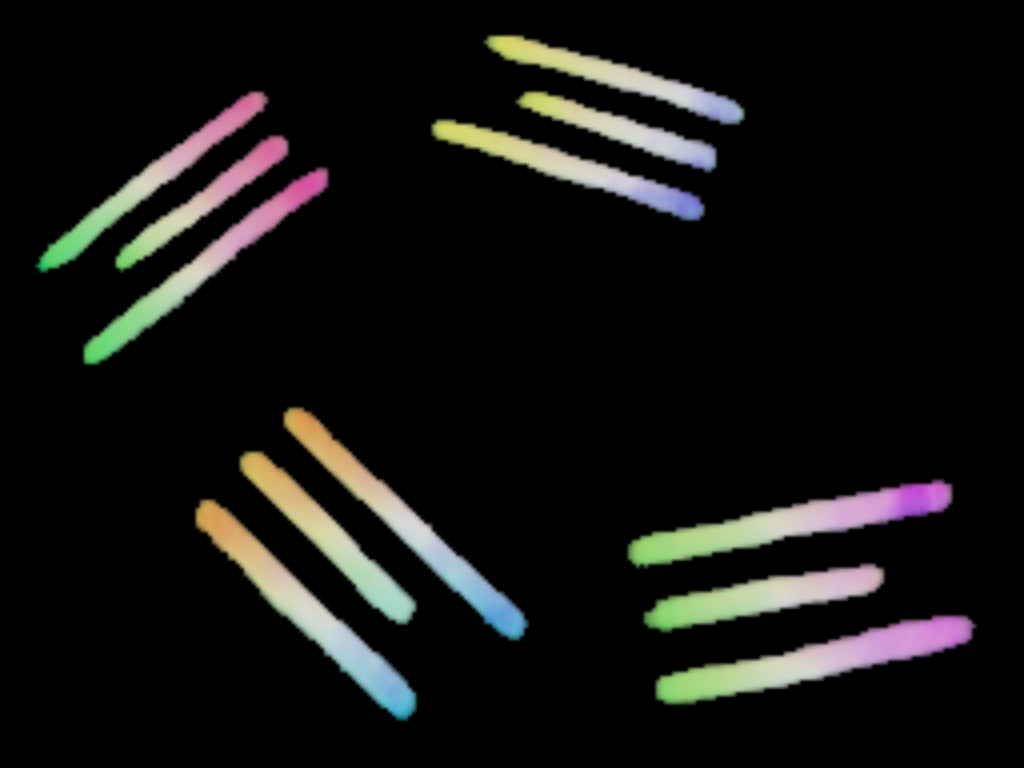}
        \centering{(a)}

		\includegraphics[width=\linewidth]{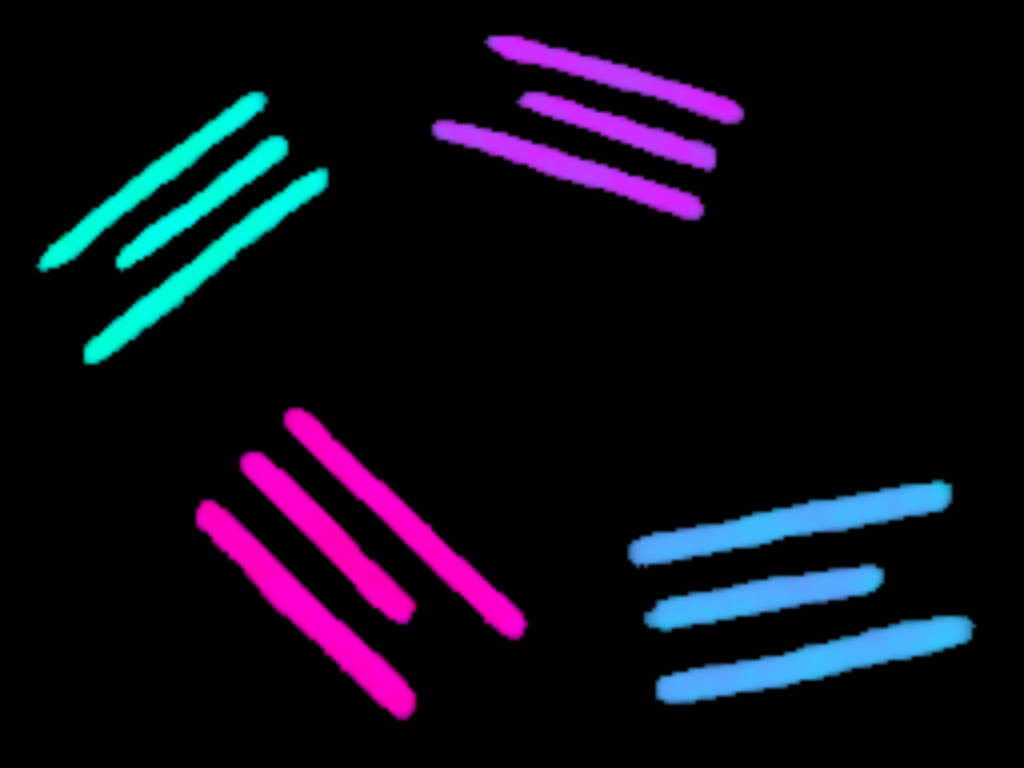}
       	\centering{(c)}

	\end{minipage}
&
	\begin{minipage}{0.45\linewidth}
		\includegraphics[width=\linewidth]{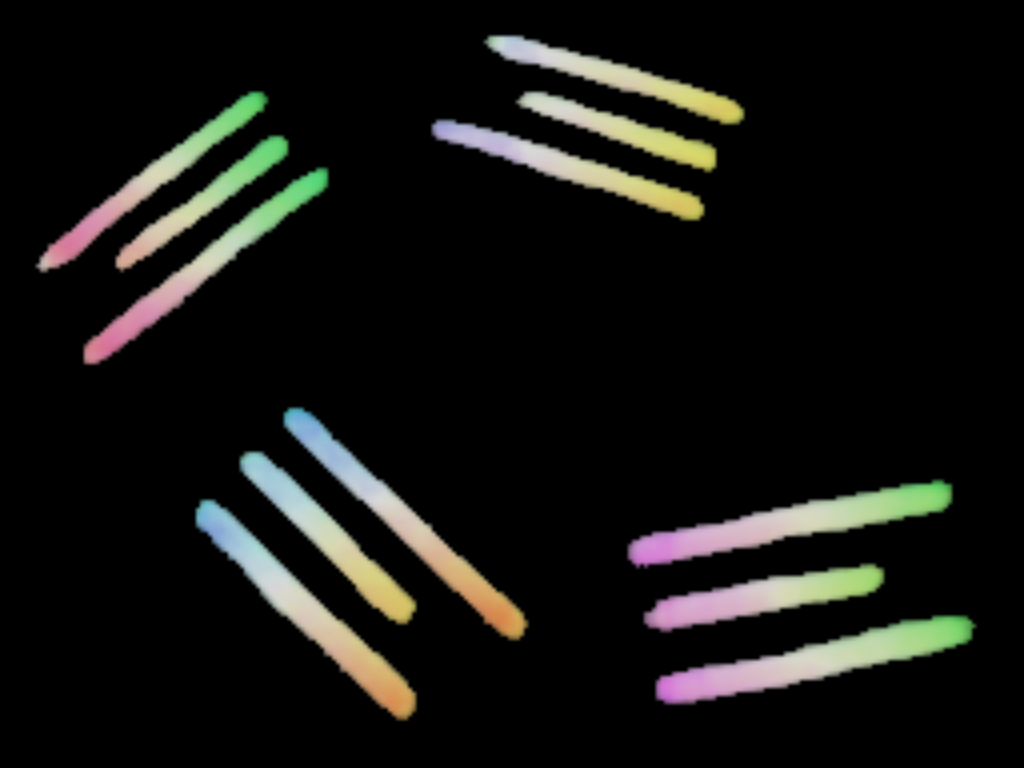}
        \centering{(b)}

		\includegraphics[width=\linewidth]{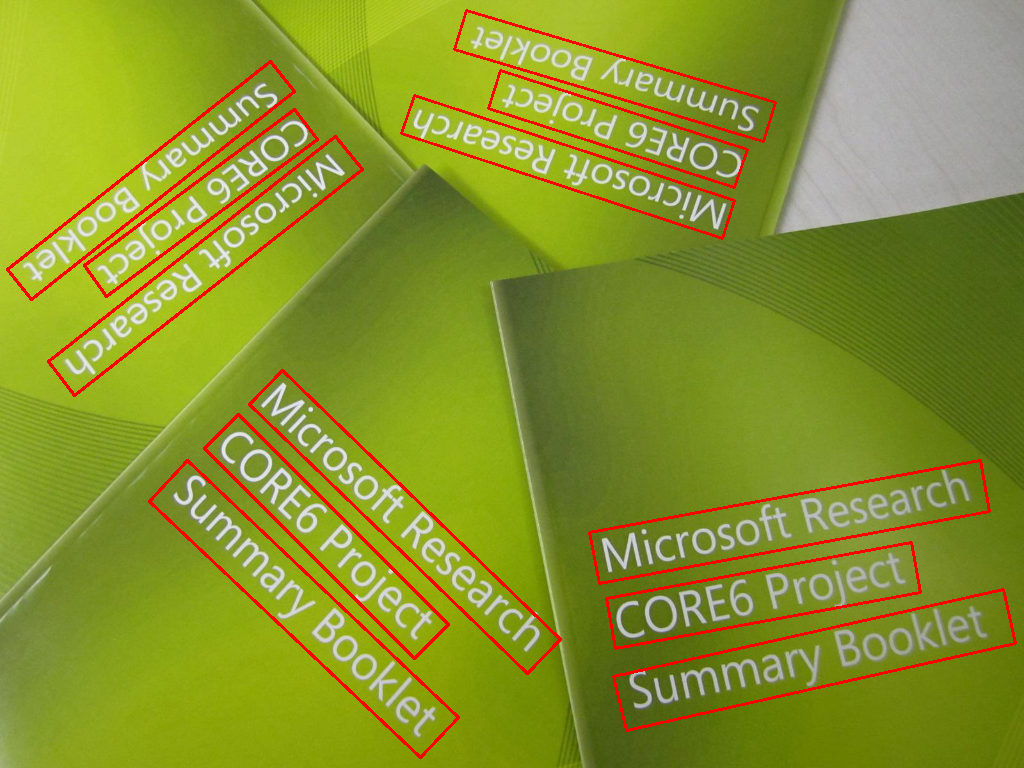}
        \centering{(d)}
	\end{minipage}
\end{tabular}
\end{center}

\vspace{-1mm}
\caption{Intermediate results of the proposed algorithm. (a) Estimated geometry map for $d_{1}$ and $d_{4}$. (b) Estimated geometry map for $d_{2}$ and $d_{3}$. (c) Estimated angle map for text instances. (d) Predicted rotated rectangles of text instances. Maps in (a), (b) and (c) are color-coded to represent variance (for $d_{1}, d_{2}, d_{3}$ and $d_{4}$) and invariance (for angle) in an pixel-wise manner. Note that in the geometry maps only the values of foreground pixels are valid.}
\label{fig:intermediate-output}
\vspace{-4mm}
\end{figure}

Fig.~\ref{fig:qualitative} depicts several detection examples by the proposed algorithm. It is able to handle various challenging scenarios, such as non-uniform illumination, low resolution, varying orientation and perspective distortion. Moreover, due to the voting mechanism in the NMS procedure, the proposed method shows a high level of stability on videos with various forms of text instances\footnote{Online video: \url{https://youtu.be/o5asMTdhmvA}. Note that each frame in the video is processed independently.}.

The intermediate results of the proposed method are illustrated in Fig.~\ref{fig:intermediate-output}. As can be seen, the trained model produces highly accurate geometry maps and score map, in which detections of text instances in varying orientations are easily formed.

\subsection{Quantitative Results}

As shown in Tab.~\ref{tab:result-icdar15} and Tab.~\ref{tab:result-coco-text}, our approach outperforms previous state-of-the-art methods by a large margin on ICDAR 2015 and COCO-Text.

In ICDAR 2015 Challenge 4, when images are fed at their original scale, the proposed method achieves an F-score of 0.7820. When tested at multiple scales \footnote{At relative scales of 0.5, 0.7, 1.0, 1.4, and 2.0.} using the same network, our method reaches 0.8072 in F-score, which is nearly 0.16  higher than the best method~\cite{yao2016scene} in terms of absolute value (0.8072 vs. 0.6477).

Comparing the results using VGG16 network\cite{tian2016detecting,Ref:Zhang2016,yao2016scene}, the proposed method also outperforms best previous work~\cite{yao2016scene} by 0.0924 when using QUAD output, 0.116 when using RBOX output. Meanwhile these networks are quite efficient, as will be shown in Sec.\ref{sec:speed}.

In COCO-Text, all of the three settings of the proposed algorithm result in higher accuracy than previous top performer~\cite{yao2016scene}. Specifically, the improvement over~\cite{yao2016scene} in F-score is 0.0614 while that in recall is 0.053, which confirm the advantage of the proposed algorithm, considering that COCO-Text is the largest and most challenging benchmark to date. Note that we also included the results from~\cite{Ref:Veit2016} as reference, but these results are actually not valid baselines, since the methods (A, B and C) are used in data annotation.

The improvements of the proposed algorithm over previous methods prove that a simple text detection pipeline, which directly targets the final goal and eliminating redundant processes, can beat elaborated pipelines, even those integrated with large neural network models.

As shown in Tab.~\ref{tab:result-msra-td500}, on MSRA-TD500  all of the three settings of our method achieve excellent results. The F-score of the best performer (Ours+PVANET2x) is slightly higher than that of~\cite{yao2016scene}. Compared with the method of Zhang~\etal~\cite{Ref:Zhang2016}, the previous published state-of-the-art system, the best performer (Ours+PVANET2x) obtains an improvement of 0.0208 in F-score and 0.0428 in precision.

Note that on MSRA-TD500 our algorithm equipped with VGG16 performs much poorer than that with PVANET and PVANET2x (0.7023 vs. 0.7445 and 0.7608), the main reason is that the effective receptive field of VGG16 is smaller than that of PVANET and PVANET2x, while the evaluation protocol of MSRA-TD500 requires text detection algorithms output line level instead of word level predictions.

\begin{table}
\small
\begin{center}
\begin{tabular}{|c|c|c|c|}
\hline
\textbf{Algorithm} & \textbf{Recall} & \textbf{Precision} & \textbf{F-score} \\\hline\hline

\begin{scriptsize} Ours + PVANET2x RBOX MS* \end{scriptsize}&
\begin{footnotesize} \textbf{0.7833} \end{footnotesize}&
\begin{footnotesize} 0.8327 \end{footnotesize}&
\begin{footnotesize} \textbf{0.8072} \end{footnotesize}\\\hline

\begin{footnotesize} Ours + PVANET2x RBOX \end{footnotesize} &
\begin{footnotesize} 0.7347 \end{footnotesize}&
\begin{footnotesize} \textbf{0.8357} \end{footnotesize}&
\begin{footnotesize} \textbf{0.7820} \end{footnotesize}\\\hline

\begin{footnotesize} Ours + PVANET2x QUAD \end{footnotesize}&
\begin{footnotesize} 0.7419 \end{footnotesize}&
\begin{footnotesize} 0.8018 \end{footnotesize}&
\begin{footnotesize} 0.7707 \end{footnotesize}\\\hline

\begin{footnotesize} Ours + VGG16 RBOX \end{footnotesize}&
\begin{footnotesize} 0.7275 \end{footnotesize}&
\begin{footnotesize} 0.8046 \end{footnotesize}&
\begin{footnotesize} 0.7641 \end{footnotesize}\\\hline

\begin{footnotesize} Ours + PVANET RBOX \end{footnotesize}&
\begin{footnotesize} 0.7135 \end{footnotesize}&
\begin{footnotesize} 0.8063 \end{footnotesize}&
\begin{footnotesize} 0.7571 \end{footnotesize}\\\hline

\begin{footnotesize} Ours + PVANET QUAD\end{footnotesize}&
\begin{footnotesize} 0.6856 \end{footnotesize}&
\begin{footnotesize} 0.8119 \end{footnotesize}&
\begin{footnotesize} 0.7434 \end{footnotesize}\\\hline

\begin{footnotesize} Ours + VGG16 QUAD \end{footnotesize}&
\begin{footnotesize} 0.6895 \end{footnotesize}&
\begin{footnotesize} 0.7987 \end{footnotesize}&
\begin{footnotesize} 0.7401 \end{footnotesize}\\\hline

\begin{footnotesize} Yao~\etal~\cite{yao2016scene} \end{footnotesize}&
\begin{footnotesize} 0.5869 \end{footnotesize}&
\begin{footnotesize} 0.7226 \end{footnotesize}&
\begin{footnotesize} 0.6477 \end{footnotesize}\\\hline
\begin{footnotesize} Tian~\etal~\cite{tian2016detecting} \end{footnotesize}&
\begin{footnotesize} 0.5156 \end{footnotesize}&
\begin{footnotesize} 0.7422 \end{footnotesize}&
\begin{footnotesize} 0.6085 \end{footnotesize}\\\hline
\begin{footnotesize}Zhang~\etal~\cite{Ref:Zhang2016} \end{footnotesize}&
\begin{footnotesize} 0.4309 \end{footnotesize}&
\begin{footnotesize} 0.7081 \end{footnotesize}&
\begin{footnotesize} 0.5358 \end{footnotesize}\\\hline
\begin{footnotesize} StradVision2~\cite{Ref:Karatzas2015} \end{footnotesize}&
\begin{footnotesize} 0.3674 \end{footnotesize}&
\begin{footnotesize} 0.7746 \end{footnotesize}&
\begin{footnotesize} 0.4984 \end{footnotesize}\\\hline
\begin{footnotesize} StradVision1~\cite{Ref:Karatzas2015} \end{footnotesize}&
\begin{footnotesize} 0.4627 \end{footnotesize}&
\begin{footnotesize} 0.5339 \end{footnotesize}&
\begin{footnotesize} 0.4957 \end{footnotesize}\\\hline
\begin{footnotesize} NJU~\cite{Ref:Karatzas2015} \end{footnotesize} &
\begin{footnotesize} 0.3625 \end{footnotesize} &
\begin{footnotesize} 0.7044 \end{footnotesize} &
\begin{footnotesize} 0.4787 \end{footnotesize} \\\hline
\begin{footnotesize} AJOU~\cite{Ref:Koo2013} \end{footnotesize}                  &
\begin{footnotesize} 0.4694 \end{footnotesize} &
\begin{footnotesize} 0.4726 \end{footnotesize} &
\begin{footnotesize} 0.4710 \end{footnotesize} \\\hline
\begin{footnotesize} Deep2Text-MO~\cite{Ref:Yin2014, Ref:Yin2015} \end{footnotesize}                  &
\begin{footnotesize} 0.3211 \end{footnotesize} &
\begin{footnotesize} 0.4959 \end{footnotesize} &
\begin{footnotesize} 0.3898 \end{footnotesize} \\\hline
\begin{footnotesize} CNN MSER~\cite{Ref:Karatzas2015} \end{footnotesize}                  &
\begin{footnotesize} 0.3442 \end{footnotesize} &
\begin{footnotesize} 0.3471 \end{footnotesize} &
\begin{footnotesize} 0.3457 \end{footnotesize} \\\hline
\end{tabular}
\end{center}

\caption{Results on ICDAR 2015 Challenge 4 Incidental Scene Text Localization task. MS means multi-scale testing.} \label{tab:result-icdar15}
\vspace{-2mm}
\end{table}

\begin{table}
\begin{center}
\begin{tabular}{|c|c|c|c|}
\hline
\textbf{Algorithm} & \textbf{Recall} & \textbf{Precision} & \textbf{F-score} \\\hline\hline

\begin{footnotesize} Ours + VGG16\end{footnotesize} &
\begin{footnotesize} \textbf{0.324} \end{footnotesize} &
\begin{footnotesize} \textbf{0.5039} \end{footnotesize} &
\begin{footnotesize} \textbf{0.3945} \end{footnotesize} \\\hline

\begin{footnotesize} Ours + PVANET2x \end{footnotesize} &
\begin{footnotesize} 0.340 \end{footnotesize} &
\begin{footnotesize} 0.406 \end{footnotesize} &
\begin{footnotesize} 0.3701 \end{footnotesize} \\\hline

\begin{footnotesize} Ours + PVANET \end{footnotesize} &
\begin{footnotesize} 0.302 \end{footnotesize} &
\begin{footnotesize} 0.3981 \end{footnotesize} &
\begin{footnotesize} 0.3424 \end{footnotesize}  \\\hline

\begin{footnotesize} Yao~\etal~\cite{yao2016scene} \end{footnotesize} &
\begin{footnotesize} 0.271 \end{footnotesize} &
\begin{footnotesize} 0.4323 \end{footnotesize} &
\begin{footnotesize} 0.3331 \end{footnotesize} \\\hline
\hline
\multicolumn{4}{|c|}{Baselines from \cite{Ref:Veit2016}} \\\hline
\begin{footnotesize} A \end{footnotesize} &
\begin{footnotesize} 0.233 \end{footnotesize} &
\begin{footnotesize} 0.8378 \end{footnotesize} &
\begin{footnotesize} 0.3648 \end{footnotesize} \\\hline
\begin{footnotesize} B \end{footnotesize} &
\begin{footnotesize} 0.107 \end{footnotesize} &
\begin{footnotesize} 0.8973 \end{footnotesize} &
\begin{footnotesize} 0.1914 \end{footnotesize} \\\hline
\begin{footnotesize} C \end{footnotesize} &
\begin{footnotesize} 0.047 \end{footnotesize} &
\begin{footnotesize} 0.1856 \end{footnotesize} &
\begin{footnotesize} 0.0747 \end{footnotesize} \\\hline

\end{tabular}
\end{center}
\caption{Results on COCO-Text.} \label{tab:result-coco-text}
\vspace{-3mm}
\end{table}

\begin{table}
\begin{center}
\begin{tabular}{|c|c|c|c|}
\hline
\textbf{Algorithm} & \textbf{Recall} & \textbf{Precision} & \textbf{F-score} \\\hline\hline

\begin{footnotesize} Ours + PVANET2x \end{footnotesize} &
\begin{footnotesize} 0.6743 \end{footnotesize} &
\begin{footnotesize} \textbf{0.8728} \end{footnotesize} &
\begin{footnotesize} \textbf{0.7608} \end{footnotesize} \\\hline

\begin{footnotesize} Ours + PVANET \end{footnotesize} &
\begin{footnotesize} 0.6713 \end{footnotesize} &
\begin{footnotesize} 0.8356 \end{footnotesize} &
\begin{footnotesize} 0.7445 \end{footnotesize} \\\hline

\begin{footnotesize} Ours + VGG16 \end{footnotesize} &
\begin{footnotesize} 0.6160 \end{footnotesize} &
\begin{footnotesize} 0.8167 \end{footnotesize} &
\begin{footnotesize} 0.7023 \end{footnotesize} \\\hline

\begin{footnotesize} Yao~\etal~\cite{yao2016scene} \end{footnotesize} &
\begin{footnotesize} \textbf{ 0.7531} \end{footnotesize} &
\begin{footnotesize} 0.7651 \end{footnotesize} &
\begin{footnotesize} 0.7591 \end{footnotesize} \\\hline
\begin{footnotesize} Zhang~\emph{et al.}~\cite{Ref:Zhang2016} \end{footnotesize}                  &
\begin{footnotesize} 0.67 \end{footnotesize} &
\begin{footnotesize} 0.83 \end{footnotesize} &
\begin{footnotesize} 0.74 \end{footnotesize} \\\hline
\begin{footnotesize} Yin~\emph{et al.}~\cite{Ref:Yin2015} \end{footnotesize}                  &
\begin{footnotesize} 0.63 \end{footnotesize} &
\begin{footnotesize} 0.81 \end{footnotesize} &
\begin{footnotesize} 0.71 \end{footnotesize} \\\hline
\begin{footnotesize} Kang~\emph{et al.}~\cite{Ref:Kang2014} \end{footnotesize}        &
\begin{footnotesize} 0.62 \end{footnotesize}  &
\begin{footnotesize} 0.71 \end{footnotesize} &
\begin{footnotesize} 0.66 \end{footnotesize} \\\hline
\begin{footnotesize} Yin~\emph{et al.}~\cite{Ref:Yin2014} \end{footnotesize}     &
\begin{footnotesize} 0.61 \end{footnotesize} &
\begin{footnotesize} 0.71 \end{footnotesize} &
\begin{footnotesize} 0.66 \end{footnotesize}\\\hline
\begin{footnotesize} TD-Mixture~\cite{Ref:Yao2012} \end{footnotesize}                         &
\begin{footnotesize} 0.63 \end{footnotesize}  &
\begin{footnotesize} 0.63 \end{footnotesize} &
\begin{footnotesize} 0.60 \end{footnotesize} \\\hline
\begin{footnotesize} TD-ICDAR~\cite{Ref:Yao2012} \end{footnotesize}   &
\begin{footnotesize} 0.52 \end{footnotesize} &
\begin{footnotesize} 0.53 \end{footnotesize} &
\begin{footnotesize} 0.50 \end{footnotesize} \\\hline
\begin{footnotesize} Epshtein~\etal~\cite{Ref:Epshtein2010} \end{footnotesize} &
\begin{footnotesize} 0.25 \end{footnotesize}  &
\begin{footnotesize} 0.25 \end{footnotesize} &
\begin{footnotesize} 0.25 \end{footnotesize} \\\hline
\end{tabular}
\end{center}
\caption{Results on MSRA-TD500.}
\label{tab:result-msra-td500}
\vspace{-3mm}
\end{table}

In addition, we also evaluated Ours+PVANET2x on the ICDAR 2013 benchmark. It achieves 0.8267, 0.9264 and 0.8737 in recall, precision and F-score, which are comparable with the previous state-of-the-art method~\cite{tian2016detecting}, which obtains 0.8298, 0.9298 and 0.8769 in recall, precision and F-score, respectively.

\subsection{Speed Comparison}\label{sec:speed}

The overall speed comparison is demonstrated in Tab.~\ref{tab:speed-comp}. The numbers we reported are averages from running through 500 test images from the ICDAR 2015 dataset at their original resolution (1280x720) using our best performing networks. These experiments were conducted on a server using a single NVIDIA Titan X graphic card with Maxwell architecture and an Intel E5-2670 v3 @ 2.30GHz CPU. For the proposed method, the post-processing includes thresholding and NMS, while others should refer to their original paper.

While the proposed method significantly outperforms state-of-the-art methods, the computation cost is kept very low, attributing to the simple and efficient pipeline. As can be observed from Tab.~\ref{tab:speed-comp}, the fastest setting of our method runs at a speed of 16.8 FPS, while slowest setting runs at 6.52 FPS. Even the best performing model Ours+PVANET2x runs at a speed of 13.2 FPS. This confirm that our method is among the most efficient text detectors that achieve state-of-the-art performance on benchmarks.

\begin{table}
\begin{center}
\small
\begin{tabular}{|c|c|c|c|c|}
\hline
\textbf{Approach} & \textbf{Res.} & \textbf{Device} & \textbf{T\textsubscript{1}/T\textsubscript{2} (ms)} & \textbf{FPS} \\\hline\hline
\begin{scriptsize} Ours + PVANET \end{scriptsize} &
\begin{footnotesize} 720p \end{footnotesize} &
\begin{footnotesize} Titan X \end{footnotesize} &
\begin{footnotesize} 58.1 / 1.5 \end{footnotesize} &
\begin{footnotesize} 16.8 \end{footnotesize}
\\\hline

\begin{scriptsize} Ours + PVANET2x \end{scriptsize} &
\begin{footnotesize} 720p \end{footnotesize} &
\begin{footnotesize} Titan X \end{footnotesize} &
\begin{footnotesize} 73.8 / 1.7 \end{footnotesize} &
\begin{footnotesize} 13.2 \end{footnotesize}
\\\hline

\begin{scriptsize} Ours + VGG16 \end{scriptsize} &
\begin{footnotesize} 720p \end{footnotesize} &
\begin{footnotesize} Titan X \end{footnotesize} &
\begin{footnotesize} 150.9 / 2.4 \end{footnotesize} &
\begin{footnotesize} 6.52 \end{footnotesize}
\\\hline

\begin{scriptsize} Yao~\etal~\cite{yao2016scene} \end{scriptsize} &
\begin{footnotesize} 480p \end{footnotesize} &
\begin{footnotesize} K40m \end{footnotesize} &
\begin{footnotesize} 420 / 200 \end{footnotesize} &
\begin{footnotesize} 1.61 \end{footnotesize}
\\\hline

\begin{scriptsize} Tian~\etal~\cite{tian2016detecting} \end{scriptsize} &
\begin{footnotesize} ss-600* \end{footnotesize} &
\begin{footnotesize} GPU \end{footnotesize} &
\begin{footnotesize} 130 / 10 \end{footnotesize} &
\begin{footnotesize} 7.14 \end{footnotesize}
\\\hline

\begin{scriptsize} Zhang~\etal~\cite{Ref:Zhang2016}* \end{scriptsize} &
\begin{footnotesize} MS* \end{footnotesize} &
\begin{footnotesize} Titan X \end{footnotesize} &
\begin{footnotesize} 2100 / N/A \end{footnotesize} &
\begin{footnotesize} 0.476 \end{footnotesize}
\\\hline

\end{tabular}
\end{center}
\caption{Overall time consumption compared on different methods. T\textsubscript{1} is the network prediction time, and T\textsubscript{2} accounts for the time used on post-processing.
For Tian~\etal~\cite{tian2016detecting}, \textit{ss-600} means short side is 600, and 130ms includes two networks. Note that they reach their best result on ICDAR 2015 using a short edge of 2000, which is much larger than ours. For Zhang~\etal~\cite{Ref:Zhang2016}, MS means they used 200, 500, 1000 three scales, and the result is obtained on MSRA-TD500. The theoretical flops per pixel for our three models are 18KOps, 44.4KOps and 331.6KOps respectively, for PVANET, PVANET2x and VGG16.
} \label{tab:speed-comp}
\vspace{-4mm}
\end{table}

\iftrue
\subsection{Limitations}

The maximal size of text instances the detector can handle is proportional to the receptive field of the network. This limits the capability of the network to predict even longer text regions like text lines running across the images.

Also, the algorithm might miss or give imprecise predictions for vertical text instances as they take only a small portion of text regions in the ICDAR 2015 training set.
\fi

\section{Conclusion and Future Work}

We have presented a scene text detector that directly produces word or line level predictions from full images with a single neural network.
By incorporating proper loss functions, the detector can predict either rotated rectangles or quadrangles for text regions, depending on specific applications. The experiments on standard benchmarks confirm that the proposed algorithm substantially outperforms previous methods in terms of both accuracy and efficiency.

Possible directions for future research include: (1) adapting the geometry formulation to allow direct detection of curved text; (2) integrating the detector with a text recognizer; (3) extending the idea to general object detection.

{\small
\bibliographystyle{ieee}
\bibliography{egbib}
}

\end{document}